\documentclass{IEEEtran}
\usepackage[utf8]{inputenc} 
\PassOptionsToPackage{hyphens}{url}\usepackage{hyperref}
\IEEEoverridecommandlockouts
\usepackage{amsmath, mathtools, amssymb}
\usepackage{multirow, multicol}
\usepackage{graphicx}
\usepackage{caption}
\usepackage{subcaption}
\usepackage{color, soul}
\usepackage{verbatim}
\usepackage{hyperref}
\usepackage{comment}
\usepackage{float}
\usepackage{cite}
\usepackage{cleveref}
\hypersetup{
  colorlinks   = true, 
  urlcolor     = blue, 
  linkcolor    = blue, 
  citecolor   = red 
}
\usepackage{xcolor}
\usepackage{blindtext}

\begin{document}
\title{ Control Modes of Teleoperated Surgical Robotic System's Tools in Ophthalmic Surgery}
\author{Haoran Wang\textsuperscript{*1}, Yasamin Foroutani\textsuperscript{*1}, Matthew Nepo\textsuperscript{1}, Mercedes Rodriguez\textsuperscript{2}, Ji Ma\textsuperscript{1}, Jean-Pierre Hubschman\textsuperscript{2}, Tsu-Chin Tsao\textsuperscript{1}, and Jacob Rosen\textsuperscript{1}

\thanks{
$^{*}$ Indicates equal contribution\\
$^{1}$Department of Mechanical and Aerospace Engineering at the University of California, Los Angeles (UCLA), USA 90095. {\tt\small \{haoranwang, yforoutani, msn629, jima, ttsao, jacobrosen\}@ucla.edu}\\
$^{2}$Jules Stein Eye Institute at the University of California, Los Angeles (UCLA), USA 90095.{\tt\small \{jphubschman,  docmercedesrodriguez\}@gmail.com}\\
}}%

\maketitle

\begin{abstract} \\ 
The introduction of a teleoperated surgical robotic system designed for minimally invasive procedures enables the emulation of two distinct control modes through a dedicated input device of the surgical console: (1) Inside Control Mode, which emulates tool manipulation near the distal end (i.e., as if the surgeon was holding the tip of the instrument  inside the patient’s body), and (2) Outside Control Mode, which emulates manipulation near the proximal end (i.e., as if the surgeon was holding the tool externally). The overarching aim of this reported research is to study and compare the surgeon’s performance utilizing these two control modes of operation along with various scaling factors in a simulated vitreoretinal surgical setting. The console of Intraocular Robotic Interventional Surgical System (IRISS)  was utilized but the surgical robot itself and the human eye anatomy was simulated by a virtual environment (VR) projected microscope view of an intraocular setup to a VR headset. Five experienced vitreoretinal surgeons and five subjects with no surgical experience used the system to perform fundamental tool/tissue tasks common to vitreoretinal surgery including: (1) touch and reset; (2) grasp and drop; (3) inject; (4) circular tracking. The results indicate that Inside Control outperforms Outside Control across multiple tasks and performance metrics. Higher scaling factors (20 and 30) generally provided better performance, particularly for reducing trajectory errors and tissue damage. This improvement suggests that larger scaling factors enable more precise control, making them the preferred option for fine manipulation tasks. However, task completion time was not consistently reduced across all conditions, indicating that surgeons may need to balance speed and accuracy/precision based on specific surgical requirements. By optimizing control dynamics and user interface, robotic teleoperation has the potential to reduce complications, enhance surgical dexterity, and expand the accessibility of high-precision procedures to a broader range of practitioners.

\end{abstract}

\begin{figure*}[!htb]
    \includegraphics[width = \textwidth]{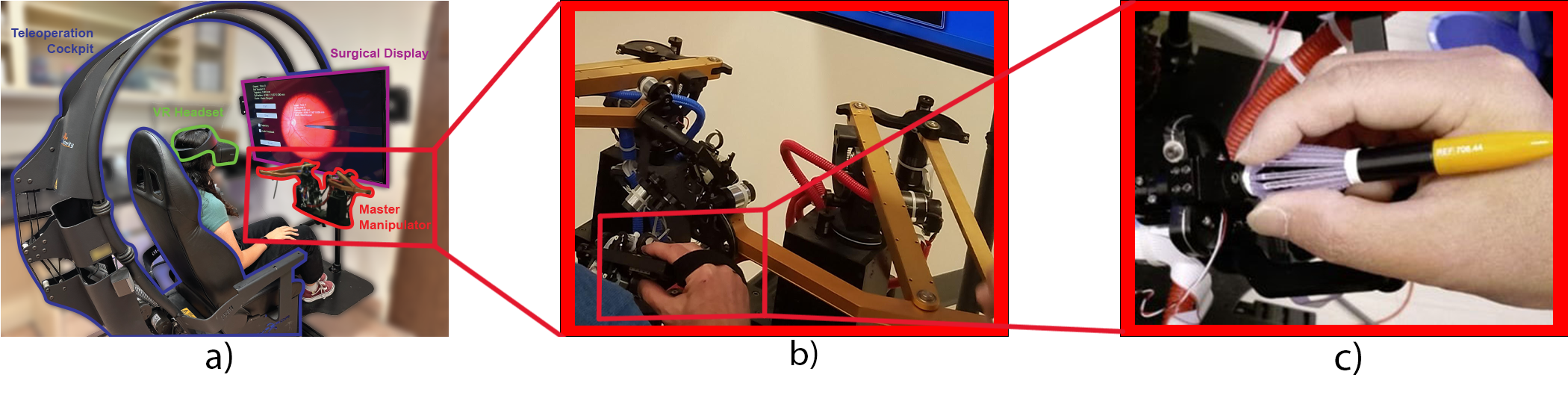}
    \caption{Experimental Teleoperation Setup (a) Master Manipulator and Cockpit (controlled by the surgeon). (b) Closeup view of Master Manipulator input device (c) Closeup view of input device Gripper Mechanism}
    \label{fig:IRISS_Setup_1}
\end{figure*}
\section{Introduction}
In Minimally Invasive Surgery (MIS), surgical instruments are introduced into the body through small ports established at the skin surface or, in the case of ophthalmic procedures, through specific ocular tissues such as the sclera, cornea, or conjunctiva. The establishment of these ports fundamentally alters the surgeon's interaction with the instrument. Unlike open surgery, where the surgeon may manipulate the tool from any position along its shaft—including proximally or distally—MIS confines the surgeon’s interaction to the proximal end of the tool, which remains external to the patient's body, while the distal end performs the intervention through the fixed port.

The presence of a port introduces several critical constraints:
(a) It reduces the number of degrees of freedom (DOF) available for tool manipulation from six to four (excluding grasping), unless wristed instruments are employed;
(b) It transforms tool motion from Cartesian space (linear and rotational movement in 3D) into a spherical motion space, pivoting around the port;
(c) It inverts the relationship between handle and tip motion, such that movement of the handle in one direction results in motion of the tool tip in the opposite direction inside the body.

To address these limitations and enhance surgical control, the use of teleoperated robotic systems in MIS enables the emulation of two distinct control modes through a dedicated input device at the surgical console:
(1) Inside Control Mode, which emulates tool manipulation near the distal end (i.e., as if the surgeon was holding the tip of the instrument  inside the patient’s body), and (2) Outside Control Mode, which emulates manipulation near the proximal end (i.e., as if the surgeon was holding the tool externally).

Vitreoretinal microsurgery is among the most challenging surgical procedures in ophthalmology, requiring high precision in extremely constrained spaces. The delicate nature of the procedure stems from the need to manipulate tissues on the micron scale, as even the largest retinal veins measure only around $150\mu m$ in diameter \cite{goldenberg2013diameters} \cite{gonenc2015force}. This intricate work demands exceptional dexterity and control, as any unintended motion can result in irreversible damage to retinal structures. Additionally, the surgical environment poses unique challenges, such as reduced spatial resolution and depth perception due to the reliance on microscopic visualization \cite{channa2017robotic}. Compounding these issues are physiological factors, including hand tremors and a limited tactile response, which further elevate the difficulty of achieving consistent surgical outcomes \cite{singh2002physiological, gupta1999surgical, patkin1967ergonomic}.
The aforementioned challenges of vitreoretinal surgery make it an optimal candidate for robotic intervention and success in addressing many of these issues has been demonstrated in the past with several robotic surgical platforms. In particular, minimization of hand tremor and improvements to surgical precision have been noted as benefits \cite{jacobsen2020robot, de2019human, maberley2020comparison}. 
Current approaches to surgery range from fully manual to fully automated robotic systems, with a spectrum of assistance levels in between. Teleoperated surgery lies within this spectrum. In this form of robot assisted surgery, the surgeon uses a microscope or a digital heads mounted display (HMD) as visual feedback while controlling the robotic system via a spatially representative control system, like input devices \cite{gerber2020advanced}.\\
An important factor when designing a teleoperated system is choosing the ideal mapping strategy and control modes. This involves choosing how the human motion will be converted to the motion of the robot's end effector. 
Various studies have been performed in this area, comparing task space and joint space mapping \cite{wang2021comparison}, as well as direct and inverse control of the robot. The overall results of these studies remain inconclusive, with some arguing that inverse control is more intuitive for surgeons \cite{zahraee2009evaluating}, others claiming direct tooltip control is the most accurate \cite{zahraee2010toward}, and occasional studies suggesting very little difference between the two control modes, with an implication that any preference is due to the amount of practice a surgeon has had with each system \cite{anderson2016comparing,tonet2006comparison}.\\
These inconsistencies become even more prominent when comparing two robotic systems currently used for human surgery. The Da Vinci Surgical System (Intuitive Surgical, Inc, Sunnyvale, CA) is currently the most prevalent commercialized, and FDA approved, teleoperated surgical system. This system directly maps the surgeon's hand position to the robot tooltip, and claims this to be the most accurate control method. While the robot is not specifically designed for eye surgery \cite{bourla2008feasibility}, it has been used to perform various ophtalmic procedures in the past \cite{tsirbas2007robotic, bourges2009robotic, hubschman2007robotic, yu1998robotic}. 

\begin{figure}
\includegraphics[width = \linewidth]{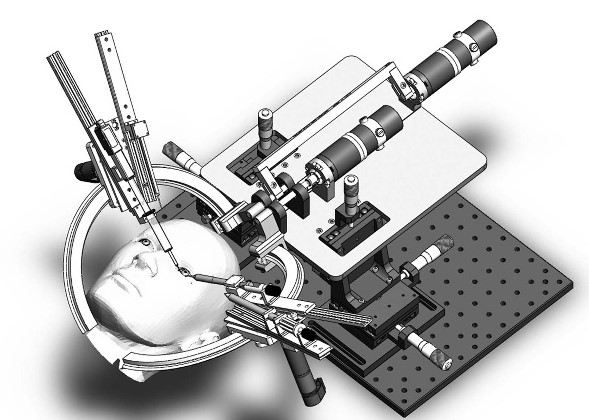}
\caption{Slave Manipulator (IRISS Robot)}
\label{fig:IRISS_Setup_2}
\end{figure}

Another system, designed more specifically for intraocular surgery, is the Preceyes Surgical System (Preceyes BV, Eindhoven, the Netherlands), which has a remote center of motion and a input device that controls the robot's movements outside of the eye, providing inverse mapping \cite{de2016robotic, edwards2018first}.
These contradictions in literature and robotic design choices can be attributed to inconsistent experimental setups, choice of test participants, and chosen surgical procedures. Additionally, no existing research specifically addresses the unique challenges of intraocular surgery in their comparisons. As such, further study is required to determine the ideal control mode for teleoperated ophthalmic surgery.\\
For the purpose of exploring these topics, this study utilized the Intraocular Robotic Intervention Surgical System (IRISS),
developed to perform teleoperated and automated intraocular surgery \cite{rahimy2013robot}. The IRISS has seen many improvements and refinements to its design, efficiency, and safety features since its initial development \cite{wilson2018intraocular}. One such effort has been the use of a surgical cockpit to determine the effect of the implementation of haptic feedback on surgical performance in epiretinal membrane peeling procedures \cite{francone2019effect}. The IRISS platform, as used in this study, is composed of the IRISS robotic system (Fig. \ref{fig:IRISS_Setup_2}) and a surgical cockpit, enabled with virtual reality display for teleoperated vitreoretinal surgery on a simulated system (Fig. \ref{fig:IRISS_Setup_1}). Using this system, two control modes are compared, which for the purpose of this study will be designated as "Inside Control" and "Outside Control." \\

\begin{figure}
    \centering
    \includegraphics[width = \linewidth]{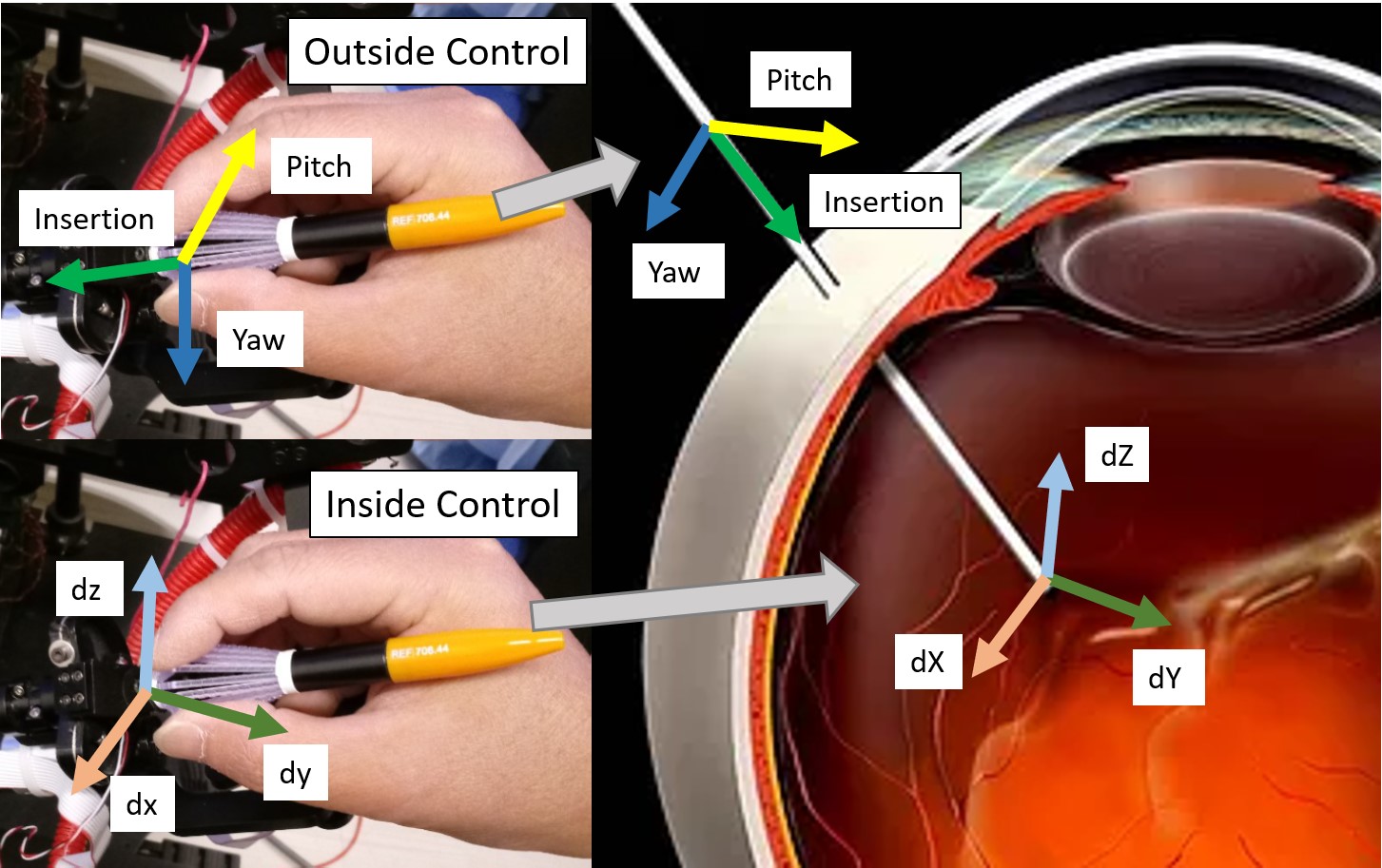}
    \caption{Coordinate frames for Inside and Outside control modes}
    \label{fig:frames}
\end{figure}

This study defines Inside Control as the direct mapping of the master tooltip's motion to the position of the slave tooltip inside the eye. For Inside Control, teleoperation directly controls the robot's tooltip position, as if the operator were simply controlling a point in space that they are able to "drive." Conversely, in Outside Control the master's movements map inversely to the motions of the slave tool, similarly to manual intraocular surgery, where the tool is held outside of the eye. In this case, the operator is manipulating a phantom tool mimicking the tools present in clinic and controls the robotic surgical device as though it were simply a more stable version of their own limb (Fig. \ref{fig:frames}).
The experimental setup divided participants into two different groups: experienced surgeons and non-surgeons (engineers).\\
To further assess the effects of mapping modes, this study also evaluates the ideal scaling factor and overall learning curves for the procedures studied. This study hypothesized that an Inside Control method may present a faster and more intuitive approach for inexperienced users, as the wrist inverted movements required in the Outside Control approach may render it more complex and time-consuming.\\

\section{Materials and Methods}
\subsection{Surgical System Design}
\begin{figure}
    \centering  
    \includegraphics[width = \linewidth]{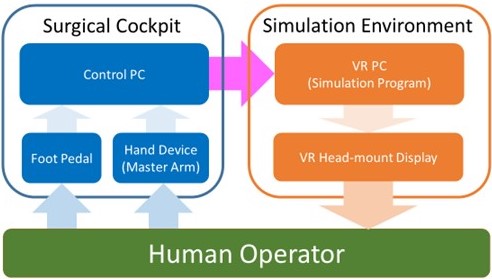}
    \caption{Teleoperation system structure}
    \label{fig:process}
\end{figure}

\begin{figure}[t]
    \centering \includegraphics[width = \linewidth]{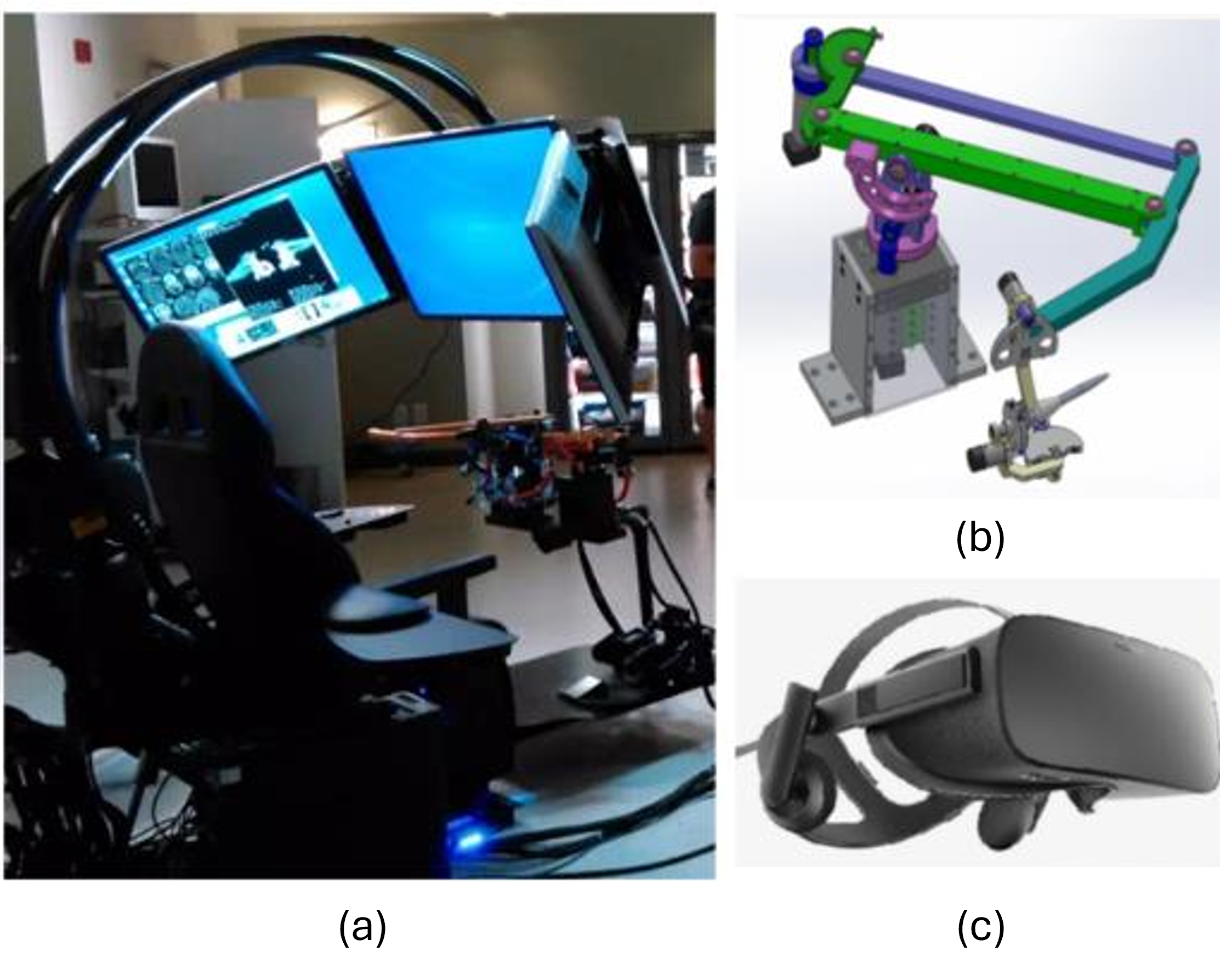}
    \caption{Experimental setup showing the surgical cockpit and master arm configuration}
    \label{cockpit}
\end{figure}

The teleoperated system is structured to replicate the intricate demands of vitreoretinal surgery, the workflow of which can be seen in Fig. \ref{fig:process}. At the core of the setup is a surgical cockpit, comprising dual master arms and a foot pedal interface for seamless operator control over the slave robotic system (Fig. \ref{cockpit}a). The master arms are engineered to offer six degrees of freedom (DOF) for precise motion (design seen in Fig.\ref{cockpit}b), allowing for translational and rotational manipulation. In addition, the gripper mechanism, controlled via an intuitive open/close DOF, replicates the functionality of microsurgical instruments typically used in ophthalmic procedures (Fig. \ref{fig:IRISS_Setup_1}c). The system prioritizes fidelity in motion replication, with algorithms compensating for inertia and latency to ensure the slave arm mimics the operator's inputs in real time. This high level of mechanical precision is critical for addressing the sub-millimeter tolerances required in intraocular surgery.

Only one of the master arms were used in this experiment. The foot pedal acts as a clutch or emergency stop, either engaging the teleoperation when activated or locking the slave arm in place when disengaged.\\
The control PC processes the input commands from the master arm, compensating for gravity and transmitting motion instructions to the slave arm. The control system ensures that the slave arm replicates the motions of the master arm with high precision, a critical requirement for delicate intraocular surgeries. The control algorithms were implemented using Microsoct VC++ and Chai3D, enabling smooth real-time interactions between the master and slave systems. This setup attempts to replicate real world surgical tasks by allowing direct control of a surgical tool similar to that used in ophthalmic surgery.
Further details about the mechanical system design and workflow can be found in \cite{francone2019effect}.

\subsection{VR System and Simulation Environment}
In this research, we aim to compare different control modes using a simulation environment, rather than controlling a slave robot, in order to create consistent experiments and quantifiable results. While controlling the system using the master arm, the operator views the simulation through an Oculus Rift S headset,
which provides a 3D replica of the retina, mimicking the microscope view surgeons have in the operating room.
The virtual reality (VR) component of the system is a central feature, designed to bridge the gap between simulated training environments and the tactile realities of intraocular surgery. The VR PC generates a highly detailed simulation environment, developed using Microsoft C\#, Blender, and Unity. This platform accurately replicates the visual and spatial challenges faced by surgeons during vitreoretinal procedures, including precise tasks such as reaching, grasping, and path following. The simulation is augmented by an Oculus Rift S headset, providing operators with an immersive 3D visual experience that closely mimics the microscope view of an actual operating room. This is an improvement over prior work that displayed the simulation on traditional 2D screens. Key surgical elements, such as the curvature of the retina and tool alignment, are rendered with sub-millimeter accuracy to ensure realism. Additionally, a simulated bleeding mechanism is incorporated to emulate the high stakes of intraocular manipulation. This visual-only feedback mechanism highlights tool penetrations through the retinal surface, mirroring real-world risks without introducing haptic or auditory distractions. Other forms of feedback were intentionally excluded to avoid distracting the operator or creating unrealistic expectations, as real-world vitreoretinal surgery provides little to no tactile feedback.

\subsection{Protocol}
A commonly used strategy in robotics is to take input in the form of Cartesian coordinates in operational space and map it to a slave arm target position in relative motion. In eye surgery, the surgical tool penetrates the eye and creates a pivot point on the ocular surface. This restricts planar motion on the surface of the eye ball, resulting in a remote center of motion (RCM) at the insertion point. This results in a 4 degree of freedom (DOF) motion for the robot. Therefore, the mapping from master arm to slave arm becomes $dx_{master} = \{dp_x^{master}, dp_y^{master}, dp_z^{master}, d\theta ^{master}\}$ into $dx_{slave} = \{dp_x^{slave}, dp_y^{slave}, dp_z^{slave}, d\theta ^{slave}\}$ where $\theta ^{master}$ and $\theta^{slave}$ are the rolling along the tool axis. In this method, the operator is able to control the position of the tool tip by moving the master arm in the same direction desired for the slave arm, as if the operator is holding the tool tip directly inside the eye ball with their own hand. We call this control method "Inside Control". However, conventional eye surgery requires the surgeon to operate the instrument by holding not the tool tip, but rather a gripping point at some position outside the eyeball from which the tool tip is spatially projected, which we call "Outside Control". This approach can be achieved by mapping the spherical coordinates from master arm operational space $dx_{master} = \{ d\Phi^{master}, d\Psi^{master}, dR^{master}, d\theta^{master}\}$ into slave arm target position. $\theta^{master}$, $\Psi^{master}$ and $R^{master}$ denote the master arm end effector’s pitch, yaw and insertion. In this case, the operator controls the tool tip position indirectly, as if holding the surgical instrument in conventional non-teleoperated eye surgery. Fig. \ref{fig:frames} shows the frames in both Inside Control mode and Outside Control mode. Although Inside Control mode has been the default in many surgical robotic systems, the Outside Control mode could present a more intuitive option for experienced ophthalmic surgeons, and therefore further testing is necessary to compare the performance between these two control methods.\\

The mapping between master arm and slave arm can greatly affect the operator’s performance. A scaling factor can be implemented to assist the surgeon in their tool handling or alter its versatility. A small scaling factor indicates agile motion for the slave arm but also makes it vulnerable to noise and disturbance, as well as more difficult to control for very small displacements. A large scaling factor, on the contrary, stabilizes the slave arm motion at the cost of increased master arm travelling distance. Therefore, finding a proper value or range of the scaling factor was also one of the goals of this research. The scaling factor was implemented as follows\\
\[dx^{master}_{4\times1} = \alpha_{4\times1}\cdot dx^{slave}_{4\times1}\]
Where $\alpha_{4\times1}$ is the scaling factor on each degree of freedom. Since rotation is involved, scaling on rotation could potentially result in disorientation of the operator over time, therefore scaling for all directions with respect to rotation was set as 1, including pitch $\Phi$ and yaw $\Psi$  for Outside Control as well as rolling $\theta$ for both control methods. Similarly, scaling on $p_x$, $p_y$ and $p_z$ was set to be the same value to avoid potential disorientation issues.

\subsection{Task Design}
\begin{figure}
    \centering
    \begin{subfigure}{\linewidth}
        \centering
        \includegraphics[width = \textwidth]{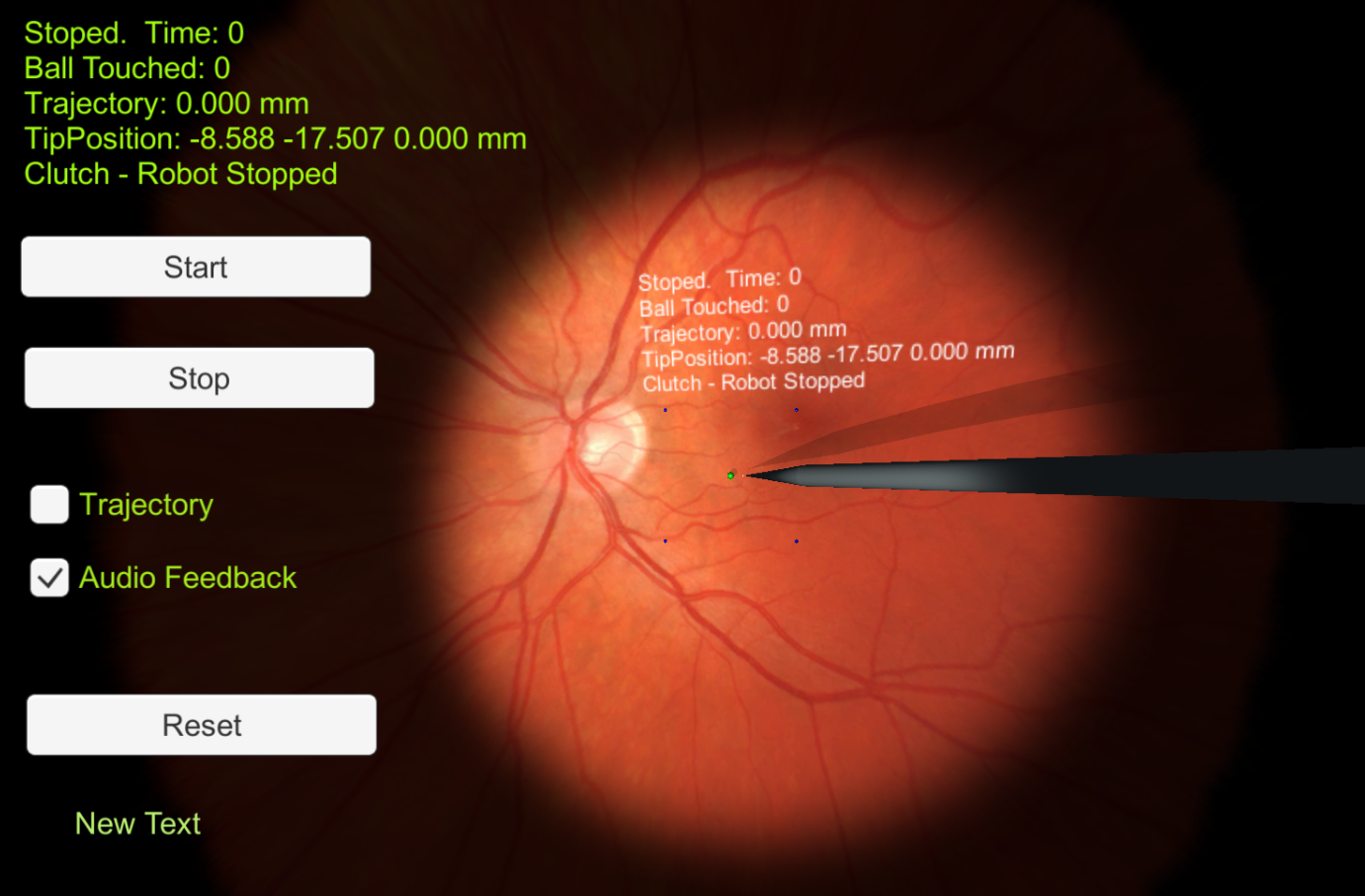}
        \caption{Surgeon view of virtual environment}
    \end{subfigure}
    \hfill
    \begin{subfigure}{0.48\linewidth}
        \centering
        \includegraphics[width = \textwidth]{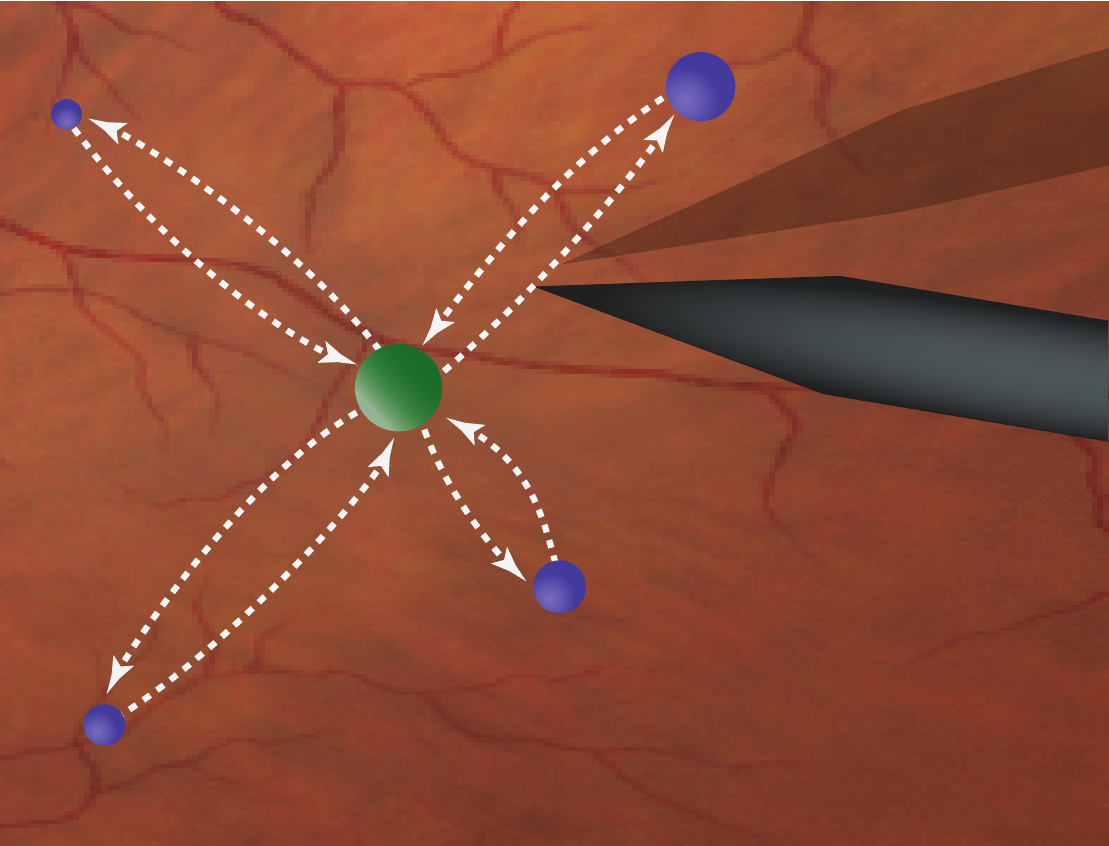}
        \caption{Task 1: Touch and reset}
        \label{task1}
    \end{subfigure}
    \hfill
    \begin{subfigure}{0.48\linewidth}
        \centering
        \includegraphics[width = \textwidth]{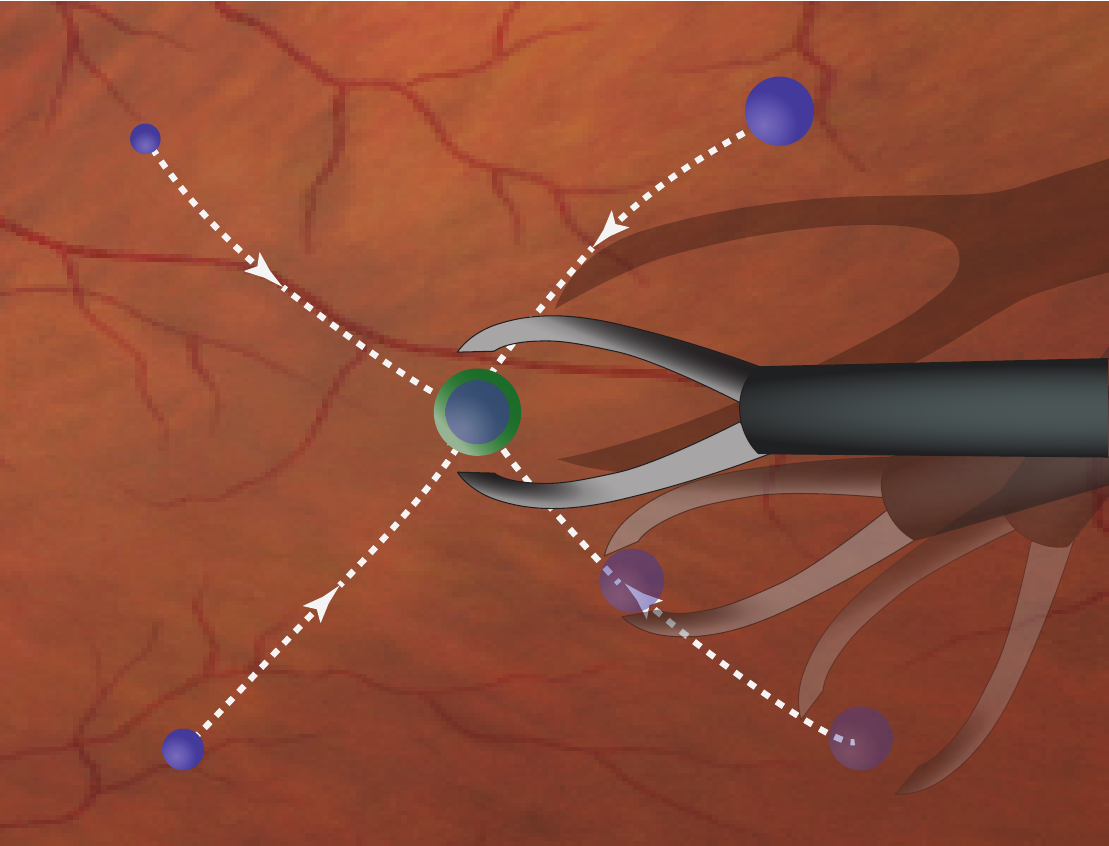}
        \caption{Task 2: Grasp and drop}
        \label{task2}
    \end{subfigure}
    \hfill
    \begin{subfigure}{0.48\linewidth}
        \centering
        \includegraphics[width = \textwidth]{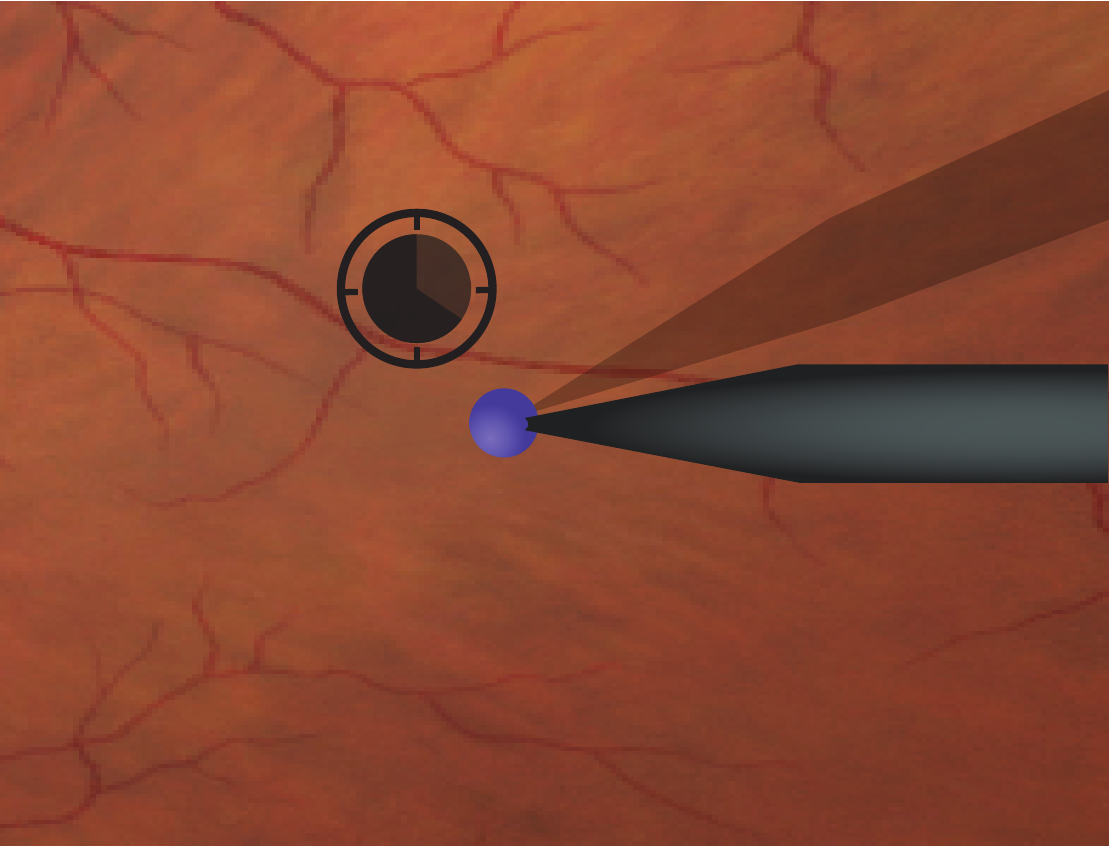}
        \caption{Task 3: Injection}
        \label{task3}
    \end{subfigure}
    \hfill
    \begin{subfigure}{0.48\linewidth}
        \centering
        \includegraphics[width = \textwidth]{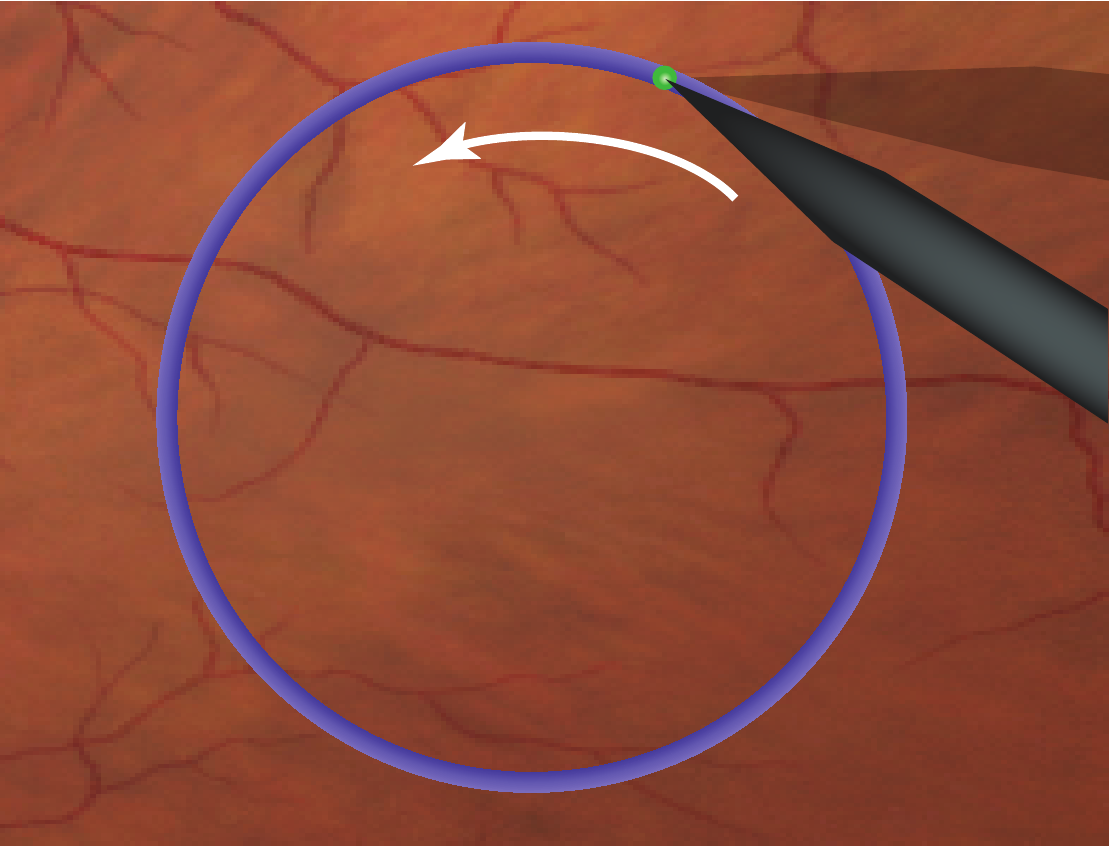}
        \caption{Task 4: Circular tracking}
        \label{task4}
    \end{subfigure}
    \caption{Experiment Tasks}
    \label{fig:tasks}
\end{figure}

In order to cover various motions that are involved during eye surgery, four tasks were designed to analyze the performance of different control settings.\\
The first task was "touch and reset", which is a basic training task to practice positioning the tool to the target. During this task, the test participant controls a needle tip and attempts to touch each of the four target spheres in clockwise order, starting with the bottom left sphere (Fig \ref{task1}). The target spheres have a diameter of $0.1mm$ and and a distance of $5.5mm$ to the task space's center. They start at a $45\deg$ angle, and occur every $90\deg s$. The tooltip starts at a "reset area" of $0.2mm$ diameter in the center, and the participant must retreat to this area after each successful touch before moving on to the next sphere. A successful touch is one where the participant is able to maintain position within the spherical area for one second. The task ends after the last successful touch of the reset area. 
To evaluate the performance, three metrics were examined; completion time, accumulated trajectory, and accumulated contact. Completion time records task time from the start within the reset area, up to the final touch of the reset area. A good performance is indicated by ability to finish the task within a limited time period when compared to manual procedures.
Accumulated trajectory is the total traveling distance of the tooltip during the task. A smaller traveling distance for the task is indicative of more stable motion and lowered chance of tissue damage. Finally, accumulated contact integrates all instances of penetration below the retinal surface, indicating the damage done to the retina during the task.\\
The second task was "grasp and drop," which is a more advanced version of the first task (Fig. \ref{task2}). In this task, the test participant needs to control a non-actuated set of forceps to reach and grasp each of the four spheres and drop them to the reset area in the same order as Task 1. The geometry and orientation of the spheres remains the same, and maintaining contact for one second is considered a successful grasp or drop. The same metrics of completion time, accumulated trajectory, and accumulated penetration were used for analysis.\\
The third task was "injection", which requires the test participant to move a needle tip to a target sphere of $0.1mm$ and maintain contact for $60$ seconds (Fig. \ref{task3}). A timer begins when the tooltip reaches the target sphere, and ends after 60 seconds. Four metrics were examined for performance in this task; accumulated trajectory, accumulated penetration, average error, and accumulated trajectory out of the target area. Average error evaluates the overall accuracy during the injection, and accumulated motion out of the target measures the total distance outside of the 0.1mm error tolerance. This measure can indicate the chance of failing by missing the target during injection.\\
The fourth task was "circular tracking", which requires the test participant to control the tip of the needle in order to drive a sphere of $0.1mm$ diameter along a circular ring with $5.5mm$ diameter. The path has a $0.15mm$ radius acceptable error, and the movement is done in the clockwise direction (Fig. \ref{task4}). The task is designed to simulate peeling operations. It starts with reaching for the target sphere, and ends when the sphere finishes one lap around the circle. Five metrics were measured in this task: completion time, accumulated trajectory, accumulated penetration, average error, and accumulated trajectory out of the tracking area.\\

\subsection{Experiment Design}
From preliminary testing, sub-millimeter level motions of the master arm were subject to delay and noise from mechanical devices, especially for translational motions. Such noise can disturb the surgeon's operation and reduce accuracy at low scaling factors. However, as the scaling factor increases, the disturbances on the slave arm also decrease. The preliminary experiments suggested that a minimum translational scaling factor of 5 is necessary to reject disturbances. The direct measurement of roll, pitch, and yaw from the motors implies that the orientation is less sensitive to disturbances. Therefore, a scaling factor of 1 can be achieved for Outside Control modes without any issue. The upper bound of the scaling factor can be found by the range of the master arm. A maximum scaling factor of 30 was found to be possible, This would allow the participant to reach the whole workspace while staying within the range of the master arm. Therefore, tests for Inside Control mode were grouped into scaling factors of 5, 10, 20, and 30. For Outside Control mode, scaling of pitch and yaw were fixed at 1 in order to avoid potential disorientation during the operation. Insertion scaling was fixed at 5, since preliminary tests suggested a large difference of scaling between orientation and insertion could also cause disorientation. In summary, 4 scaling factors for Inside Control, and 1 for Outside Control, were tested on each of the 4 tasks. It is worth noting that this study acknowledges the testing of multiple scaling factors of Inside Control for each task results in the appearance of a greater sample size for Inside Control, or an increased likelihood of participant familiarity with the Inside Control mode. However, scaling factor greatly influences handling of the system, as can be seen further in the results section, and as such each individual scaling factor of the Inside Control mode can be considered a unique category for statistical analysis.\
Another common issue that can affect the surgeons' performance besides control mode and scaling factor is the learning curve of tasks. To avoid insufficient training or overtraining throughout the tasks, all participants were given one minute of practice. During practice, each participant was allowed use of the clutch or arm rest to adjust the master arm to the most comfortable initial position for them. Clutch function was not allowed during the tasks. All participants performed the 4 tasks using the designed control modes (20 trials each), but the order was randomized for each test participant so that the data would not indicate overtraining for certain tasks. Motion data was collected from each participant, and the metrics for the tasks were calculated based on the collected data after post processing.\\

\subsection{Participants}
Ten participants were enrolled in the study: five ophthalmologists with vitreoretinal surgical training (experienced surgeons) and five engineers with no surgical experience (non-surgeons). All participants were at least 18 years of age, possessed normal sensory and motor function in their arms and hands, and had normal or corrected-to-normal visual acuity. All participants were presented with a detailed information sheet on what the study entailed and all expressed verbal consent to participate. The study protocol received approval from the University of California Los Angeles Institutional Review Board (IRB \#20-001861) and adhered to the tenets of the Declaration of Helsinki.\\

\section{Results}
\begin{figure}
    \includegraphics[width = 0.45\textwidth]{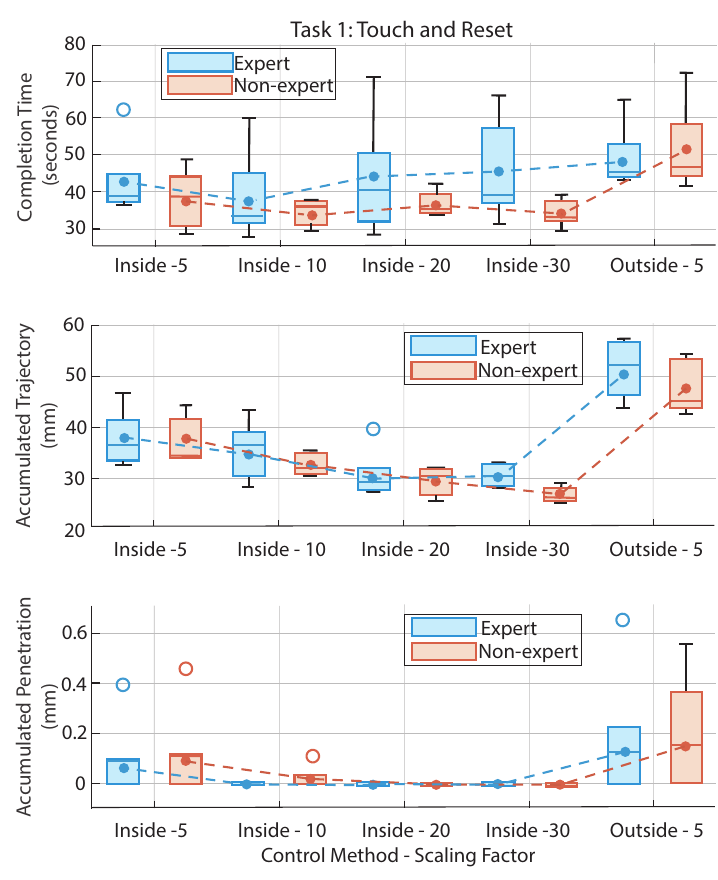}
    \caption{Performance for Task 1: Touch and reset }
    \label{performance1}
\end{figure}

In this study, all ten participants perform the same task under five distinct control conditions: four Inside control modes of varying scaling factors and one Outside control mode, for which the scaling factor must be fixed at the chosen value of 5. Each participant experiences all five control configurations, allowing for within-subject comparisons. Thus, this experimental setup constitutes a repeated measures design \cite{sheldon1996use}. Although some of these conditions involve different scaling parameters, they represent discrete and independent control strategies that are not directly comparable across modes, e.g., a scaling factor of 5 under Inside control does not correspond to a meaningful equivalent in Outside control. Therefore, it is inappropriate to treat the data as coming from a two-factor design. Instead, each of the five control schemes is considered a distinct level of a single factor, and the analysis is performed using a one-way repeated measures framework. This is in line with the suggested statistical procedure for comparison of experimental independent variables that are unique and distinct but may be affected by subject characteristics (i.e., the effect of expertise). In these cases, a single-factor, one-way, repeated measures framework is sufficient, so long as some level of analysis is done to acknowledge possible correlation, as is performed in this study with the analysis of the expert group \cite{sullivan2008repeated}.\\
Given our study’s limited sample size of ten participants, we cannot confidently assume normal distribution or homogeneity of variances across the data. To assess these assumptions, we conducted the Shapiro-Wilk test for normality and Levene’s test for equality of variances. These tests are particularly important in small sample studies as parametric tests, such as the commonly used t-test and ANOVA, are not robust to deviations from a Gaussian distribution. The results indicate that for most conditions, these assumptions were violated, confirming the necessity of non-parametric statistical methods which do not require a Gaussian distribution and are more suitable for small sample sizes \cite{pett2015nonparametric}.\\
To analyze differences between the control conditions, the Friedman test, a non-parametric alternative to the repeated measures ANOVA test, is selected to assess whether significant differences exist across the control modes. If significance was detected, we follow it with a post-hoc Wilcoxon Signed-Rank test for pairwise comparisons. Bonferroni correction was applied to control the family-wise error rate due to multiple comparisons. Cliff’s Delta $\delta$ effect sizes were also computed for all significant differences to assess their practical significance. The comparison is done first between each inside control mode and outside control, and then pairwise among all scaling factors of inside control. This approach aligns with best practices recommended for small sample analyses in use studies.\cite{pereira2015overview, pett2015nonparametric}\\
To evaluate whether prior surgical experience influenced performance, we first compared expert and non-expert participant results across all tasks. The Mann-Whitney U test was performed and revealed no significant differences in performance in most conditions, except for Task 1 and 3, where non-experts showed significantly higher out of target trajectory errors. Given the general similarity in performance, we combined data from both groups to increase statistical power in control mode comparisons. \\
Friedman tests indicated significant differences among control modes across all tasks ($p<0.001$), therefore, we performed pairwise Wilcoxon Signed-Rank tests to determine specific differences in all parameters.\\
For Task 1 (Fig. \ref{performance1}), in terms of completion time, Inside Control with a scaling factor of 10 (Inside 10) was significantly faster than Outside Control ($p=0.025,\delta=1.17$), with Inside 20 and Inside 30 also showing improvements ($p\approx 0.05,\delta \approx 0.63$). No significant differences were found among Inside Control scaling factors. When analyzing trajectory, all Inside Control modes significantly outperformed Outside Control ($p<0.001,\delta =2.38 $ to $8.46$), with Inside 20 and 30 again performing the best. Similarly, Inside 20 and 30 demonstrated superior precision resulting in almost no retinal penetration ($p=0.003$), further highlighting their suitability for delicate surgical maneuvers.\\

\begin{figure}
    \includegraphics[width = 0.45\textwidth]{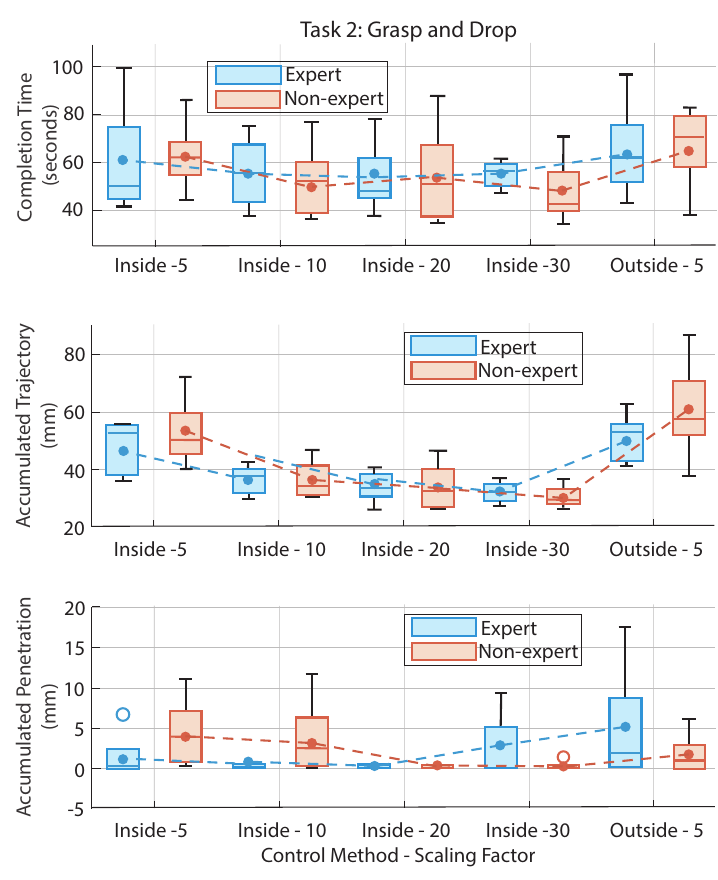}
    \caption{Performance for Task 2: Grasp and drop}
    \label{performance2}
\end{figure}

For Task 2 (Fig. \ref{performance2}), all Inside Control modes showed significantly faster completion times than Outside Control while also resulting in shorter trajectories ($p<0.001$ for all conditions, with $\delta=3.14$ to $8.79$). Unlike in Task 1, a scaling factor of 30 outperformed 5 in both time ($p=0.02, \delta=1.06$) and trajectory ($p<0.001,\delta=5.95$), suggesting that for this task larger scaling factors allow faster task execution without reducing movement accuracy. As expected with the increased difficulty of the task, it exhibited higher retinal penetration compared to Task 1, though Inside 20 still demonstrated almost no penetration, with a significant difference compared to Outside Control ($p=0.024$).\\
 
\begin{figure}
    \includegraphics[width = 0.45\textwidth]{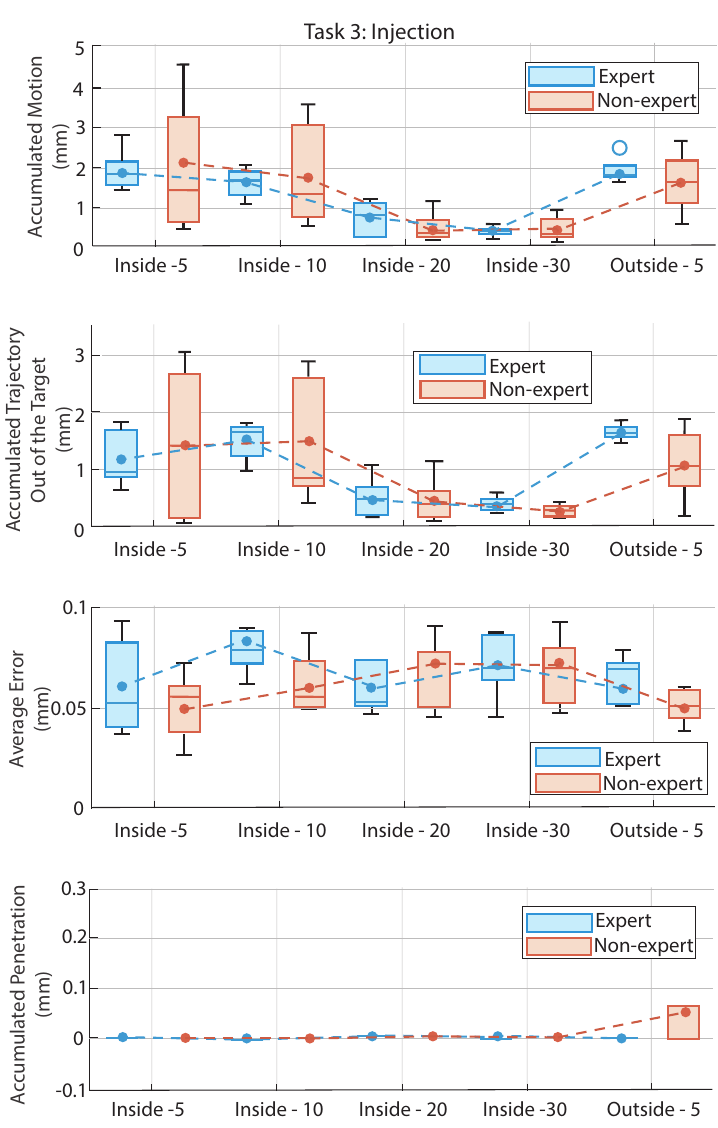}
    \caption{Performance for Task 3: njection}
    \label{performance3}
\end{figure}

For Task 3 (Fig. \ref{performance3}), Inside control at scaling factors of 20 and 30 showed significantly better performance in total motion ($p=0.01$ and $p=0.02$ respectively). These scaling factors also outperformed the other Inside Control modes in pairwise comparison, while no significant differences were observed between the two. However, when comparing average error and retinal penetration, no significant differences were observed among the conditions, as all participants successfully performed the tasks with minimal errors.\\

\begin{figure}
    \includegraphics[width = 0.45\textwidth]{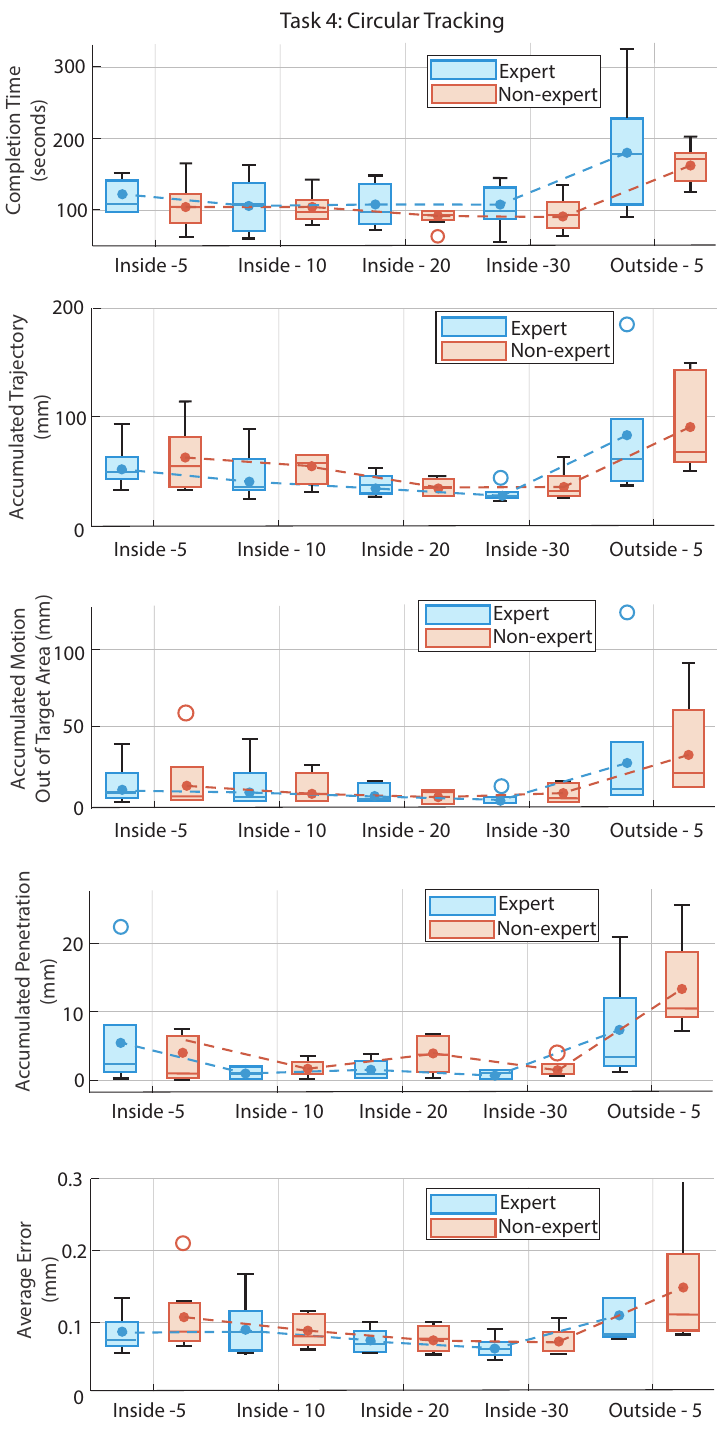}
    \caption{Performance for Task 4: Circular tracking}
    \label{performance4}
\end{figure}

\begin{figure}[t]
    \centering
    \includegraphics[width = \linewidth]{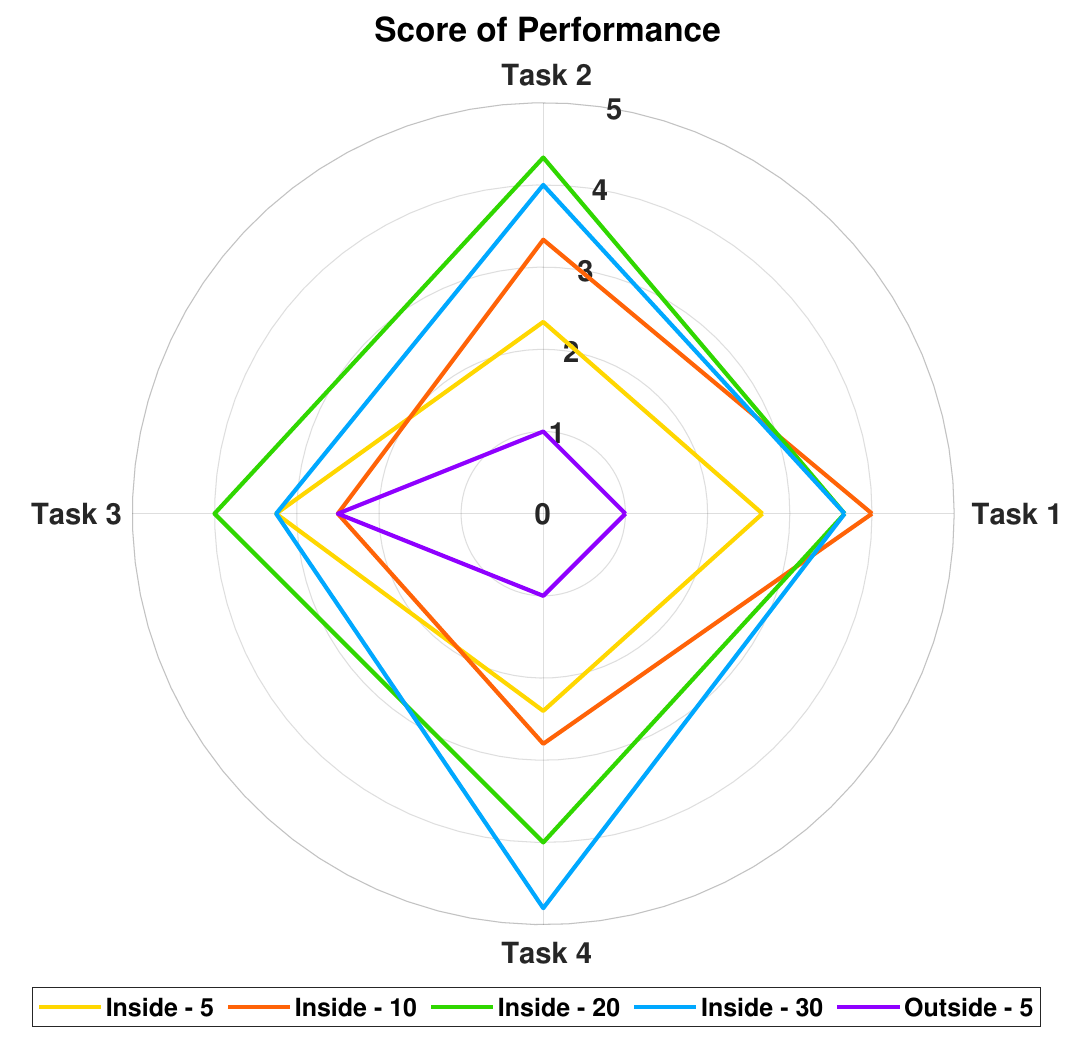}
    \caption{Score of Performance (higher is better)}
    \label{score}
\end{figure}

For Task 4 (Fig. \ref{performance4}), while completion time did not show a significant difference among different Inside Control modes, the scaling factors of 10, 20, and 30 all outperformed Outside Control ($p$-values ranging from $0.01$ to $0.03$). A similar trend was observed in the total retinal penetration in this task (p-values from 0.004 to 0.04). In trajectory and error metrics, only Inside 20 and 30 showed significant improvements (p-values ranging from 0.001 to 0.02 for the metrics). However, after applying Bonferroni correction, none of the parameters showed significant differences between the Inside Control scaling factors for this task.\\
\\ 
In addition to the statistical analysis on the full dataset, examining the results of only the experienced participants can help validate the main findings. To facilitate comparison across different scaling factors, we generated a radar chart summarizing the performance scores for each control configuration across all four tasks, using data from the five experienced surgeons (Fig. \ref{score}). This visualization provides a more integrated view of task performance and offers an intuitive way to assess the effects of the various scaling factors. The scoring criteria were as follows:
\begin{itemize}
    \item For each metric, the best performance gets 5 points, the second best gets 4 points, etc. 
    \item Performance is ranked by mean value. For the same mean value, a smaller standard deviation is considered to be better.
    \item If both mean value and standard deviation are the same, they will be tied at the same point.
    \item Total score of each scaling factor in each task is scaled to 10 for ease of reading. 
\end{itemize}

The results indicate that Inside Control outperforms Outside Control across multiple tasks and performance metrics. Higher scaling factors (20 and 30) generally provided better performance, particularly for reducing trajectory errors and tissue damage. This improvement suggests that larger scaling factors enable more precise control, making them the preferred option for fine manipulation tasks. However, task completion time was not consistently reduced across all conditions, indicating that surgeons may need to balance speed and precision based on specific surgical requirements. 
The radar chart visualization further supports these findings, as Inside 20 and 30 exhibited the highest overall scores in most tasks, followed by 10 and 5. We see, however, that for simpler tasks such as Task 1, a scaling factor of 10 can provide an optimal balance between time and precision. This is especially pronounced in the experienced surgeons who are better prepared for such tasks. Even in this category though, higher scaling factors demonstrated superior control and accuracy in the more complex procedures, particularly in avoiding errors when performing precise movements such as operating in the limited ring-shaped region of Task 4. These findings emphasize the importance of customized scaling factor adjustment tailored to the complexity and intricacies of each surgical procedure and highlight the importance of implementing more advanced adaptive strategies for real time scaling factor adjustment. In doing so, we can significantly enhance precision and safety in robotic surgery.\\
A key consideration in the teleoperation design is the surgeon's confidence in the system and ability to intervene directly, particularly given that surgeons are trained to operate in immediate proximity to the patient's head. The introduction of a surgical cockpit away from the patient raises concerns regarding awareness and response time in case of complications. However, feedback from clinicians in this study suggests that the effectiveness of the teleoperation system depends primarily on the availability and reliability of real-time feedback mechanisms rather than physical proximity. Specifically, the addition of no-fly zones, image guidance, proximity alerts, and other safety warnings provide the necessary situational awareness to make the system comfortable and secure. While direct hands-on intervention allows for immediate action, a well-designed feedback system that alerts the surgeon and can stop the procedure before adverse events occur can be just as, if not more effective in ensuring safety. As such, future work should also focus on refining these live feedback mechanisms, enhancing the safety and reliability of teleoperation.\\ 

\section{Conclusions}
Robotic teleoperation in ophthalmology remains a critical area of research, offering the potential to enhance surgical precision and patient outcomes. This study provides key insights into achieving safe and effective teleoperated surgery. The results strongly support the advantages of the Inside Control mode for teleoperated ophthalmic procedures, demonstrating superior precision, stability, and ease of use across all evaluated tasks. These benefits were particularly pronounced when Inside Control was paired with larger scaling factors, which enable fine-tuned movements critical for high-complexity surgical scenarios within confined environments. Both surgeons and non-surgeons found Inside Control straightforward and intuitive to use, which is evident through the lack of significant differences between the outcomes of the two groups. While Outside Control was initially assumed to be more familiar to experienced surgeons due to its alignment with manual surgery techniques, the results did not show a performance advantage. Instead, its higher penetration rates suggest increased surgical risk, particularly in tasks requiring high precision. Despite the conventional use of Outside Control in the majority of robotic systems designed for ophthalmology, its consistent outperformance in this study highlights the need for a paradigm shift in system design and training approaches.\\
The radar chart analysis further reinforces these findings, illustrating that scaling factors of 20 and 30 consistently outperform the others across various metrics. These results suggest that higher scaling factors allow for greater precision while maintaining control, making them an ideal choice for delicate, high-accuracy procedures. Conversely, lower scaling factors may still be advantageous for tasks that require faster tool motion and broader maneuverability. These findings emphasize the importance of tailoring the choice of control mode and scaling factor to the specific demands of different surgical tasks, enhancing the adaptability and performance of robotic systems. \\
A limitation of this study is the imbalance in trial distribution, with more trials conducted using Inside Control. This was unavoidable due to the incompatibility of a higher scaling factor with Outside Control, as preliminary testing revealed that values greater than 5 induced disorientation and compromised usability. However, several measures were implemented to minimize potential biases, including randomized trial orders and pre-trial practice sessions for all subjects. Given that the study was designed as a preliminary assessment of safety and intuitiveness, it was conducted with a limited number of participants and restricted prior training time.\\

There are two key ways in which minimally invasive (MIS) ophthalmic surgery differs from MIS procedures in general when utilizing teleoperated surgical robotic systems. These differences have important implications for the control mode paradigms. (1) Tool Control Responsibility and Workflow: In ophthalmic surgery, the surgeon maintains continuous control of the instrument via the teleoperation console before, during, and after insertion into the eye. The surgical assistant (nurse) mounts the instrument onto the robotic arm, but the surgeon actively controls the tool throughout its entire trajectory, including the critical phase of insertion through the ocular port (e.g., the sclera or cornea). This contrasts with general MIS procedures, where the instrument is typically inserted into the patient by the assistant until the tip is visible in the endoscopic field of view. Only then does the surgeon assume control via the input device. As a result, ophthalmic surgery requires a more comprehensive control strategy, and often relies on an extended visual field, such as a wide-angle digital microscope view that includes both intraocular and extraocular portions of the instrument. (2) Spatial Relationship Between Port and Surgical Site: In ophthalmic surgery, the surgical site is located very close to the port of entry, and both the intraocular and extraocular portions of the tool are often visible in the surgical microscope’s field of view. In contrast, general MIS typically involves longer instruments, where the operative site is located deeper within the body and far from the entry port. The external portion of the tool remains outside the camera’s view, as the endoscope is focused tightly on the internal surgical field. This spatial difference affects how the surgeon perceives tool motion and may influence the control mode preferences. It is important to note that these unique characteristics of ophthalmic MIS were not directly investigated in the present study. However, they may significantly influence the choice of control mode—specifically between inside and outside Control Modes. For instance, cataract surgery, where the anatomical target is very close to the incision, may favor one control mode, whereas vitreoretinal procedures (simulated in this study), where the target is located deeper within the eye, may benefit from a different mode.

Future research will focus on expanding the participant pool to enable more robust statistical analyses, improving the generalizability of the findings. Additionally, investigating the effects of 3D visualization of a curved retina and task performance in hard-to-reach regions of the intraocular space could provide valuable insights. A longitudinal learning curve study could assess how proficiency with Inside and Outside Control evolves over time, quantifying both initial adaptation speed and long-term mastery. Furthermore, adaptive scaling factors that dynamically adjust based on task complexity or real-time force feedback present promising avenues for further study. \\
Finally, validating these findings in a clinical environment will be crucial for bridging the gap between simulation and real-world applications. As the field of robotic teleoperation continues to evolve, these findings provide a roadmap for refining robotic teleoperation, ultimately advancing the efficacy and safety of ophthalmic surgery. By optimizing control dynamics and user interface, robotic teleoperation has the potential to reduce complications, enhance surgical dexterity, and expand the accessibility of high-precision procedures to a broader range of practitioners.

\bibliographystyle{IEEEtran}
\bibliography{references}
\end{document}